\documentclass[letterpaper, 10 pt, conference]{ieeeconf}  

\IEEEoverridecommandlockouts                              

\makeatletter
\let\NAT@parse\undefined
\makeatother
\usepackage[numbers,sort&compress]{natbib}

\usepackage[pdftex]{graphicx}
\usepackage{amsmath}
\usepackage{amssymb}  
\usepackage{multirow}
\usepackage{array,booktabs}
\usepackage{diagbox}
\usepackage{balance}
\usepackage{xspace}
\usepackage{pifont}
\usepackage{algorithm}
\usepackage{algpseudocode}
\usepackage{adjustbox}
\usepackage{romannum}
\usepackage{caption}
\usepackage{subcaption}
\usepackage[final]{hyperref}
\hypersetup{
 colorlinks=true,
 linkcolor=blue,
 filecolor=magenta,
 urlcolor=blue,
 citecolor=black
}
\usepackage{cleveref}
\crefname{figure}{Fig.}{Figs.}
\Crefname{figure}{Fig.}{Figs.}
\usepackage[font=small]{caption}
\usepackage{rpm_SIunits}
\usepackage{rpm_acronyms}
\usepackage{rpm_math}
\usepackage{rpm_misc}

\usepackage{soul,color}
\usepackage{lipsum}

\usepackage{xcolor}

\usepackage{tikz} 
\usepackage{subcaption}
\usepackage{caption}
\title{\LARGE \bf TranSplat: Surface Embedding-guided 3D Gaussian Splatting for Transparent Object Manipulation}     

\author{Jeongyun Kim${}^{1}$, Jeongho Noh${}^{1}$, Dong-Guw Lee${}^{1}$  and Ayoung Kim${}^{1*}$
\thanks{$^{1}$J. Kim, J. Noh, D. Lee and A. Kim are with the Dept. of Mechanical Engineering, SNU, Seoul, S. Korea {\tt\small [jeongyun, shwjdgh3842, donkeymouse, ayoungk]@snu.ac.kr}}%
}

\begin{document}

\maketitle
\thispagestyle{empty}
\pagestyle{empty}

\begin{abstract}

Transparent object manipulation remains a significant challenge in robotics due to the difficulty of acquiring accurate and dense depth measurements. Conventional depth sensors often fail with transparent objects, resulting in incomplete or erroneous depth data. Existing depth completion methods struggle with interframe consistency and incorrectly model transparent objects as Lambertian surfaces, leading to poor depth reconstruction. To address these challenges, we propose TranSplat, a surface embedding-guided 3D Gaussian Splatting method tailored for transparent objects. TranSplat uses a latent diffusion model to generate surface embeddings that provide consistent and continuous representations, making it robust to changes in viewpoint and lighting. By integrating these surface embeddings with input RGB images, TranSplat effectively captures the complexities of transparent surfaces, enhancing the splatting of 3D Gaussians and improving depth completion. Evaluations on synthetic and real-world transparent object benchmarks, as well as robot grasping tasks, show that TranSplat achieves accurate and dense depth completion, demonstrating its effectiveness in practical applications. We open-source synthetic dataset and model: \url{https://github.com/jeongyun0609/TranSplat}

\end{abstract}
\section{Introduction}
\label{sec:intro}



Manipulating transparent objects is a significant challenge in robotics, as standard depth sensors and depth completion methods often fail to provide accurate reconstructions due to the reflections and refractions inherent in transparent materials. These optical phenomena result in incomplete depth maps, noise, and artifacts, leading to incorrect 3D perception and errors in estimating grasping points.

To address these challenges, previous solutions have focused on hardware or learning-based approaches. Hardware-based methods use additional sensors, such as thermal infrared cameras \cite{huo-2023-tip} or polarized cameras \cite{mei-2022-cvpr, kalra-2020-cvpr}, to provide auxiliary depth information. However, thermal cameras are costly to operate, and polarized cameras require specific polarized cues, complicating hardware setups.

More recent solutions emphasize learning-based methods for depth completion using single-view \cite{zhu-2021-cvpr, fang-2022-ral} and multi-view RGB images \cite{kerr-2022-corl, ichnowski-2022-corl, duisterhof2024residualnerf}, facilitated by datasets specifically targeting transparent objects \cite{wang-2022-cvpr, bashkirova-2022-cvpr, chen-2022-eccv}. Multi-view RGB methods, particularly those leveraging \ac{NeRF}, offer more robust depth completion by improving occlusion handling and scale consistency. However, existing methods face three key limitations. First, transparent objects, as non-Lambertian surfaces \cite{sajjan2020clear}, are highly sensitive to changes in illumination and viewpoint, causing photometric inconsistencies. When using NeRF or \ac{3D-GS} for depth rendering, these inconsistencies introduce noise and artifacts in the depth maps. Second, directly rendering transparent surfaces based solely on RGB images often causes opacity values to collapse to zero \cite{lee2023nfl}, resulting in holes in the reconstructed depth. Third, NeRF-based techniques for novel view synthesis of transparent objects, despite recent advancements \cite{ichnowski-2022-corl,duisterhof2024residualnerf}, still suffer from slow inference times.


\begin{figure}[!t]
    \centering
    \includegraphics[width=1\linewidth]{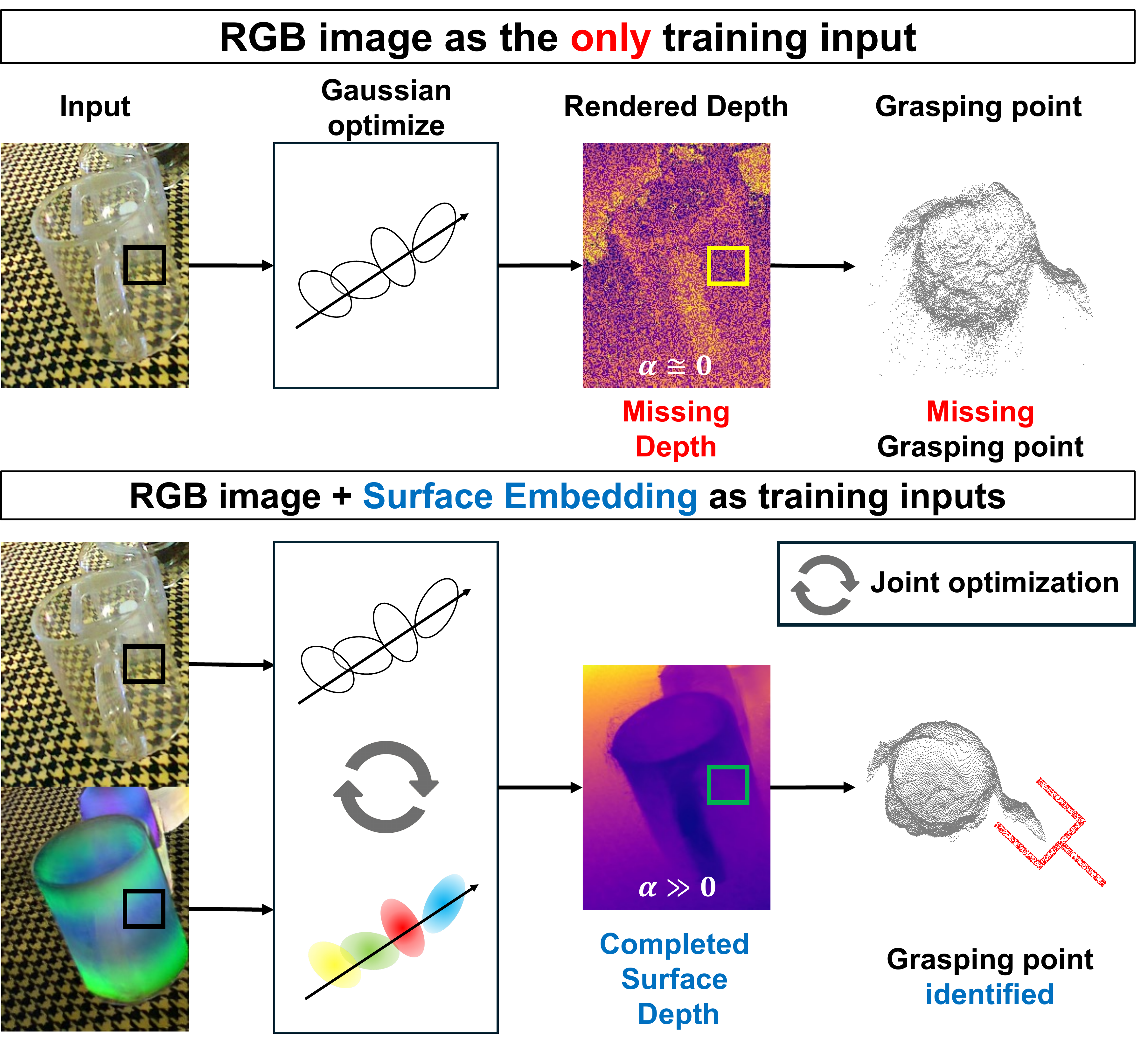}
    \caption{TranSplat optimizes 3D Gaussian splatting by jointly training with RGB and surface embeddings as inputs. This approach prevents the opacity of transparent objects from collapsing to zero and ensures smooth rendering, leading to accurate depth completion and reliable grasping points.}
    \label{fig:figure1}
    \vspace{-7mm}
\end{figure}


\begin{figure*}[t]
    \centering
    \includegraphics[width=\linewidth]{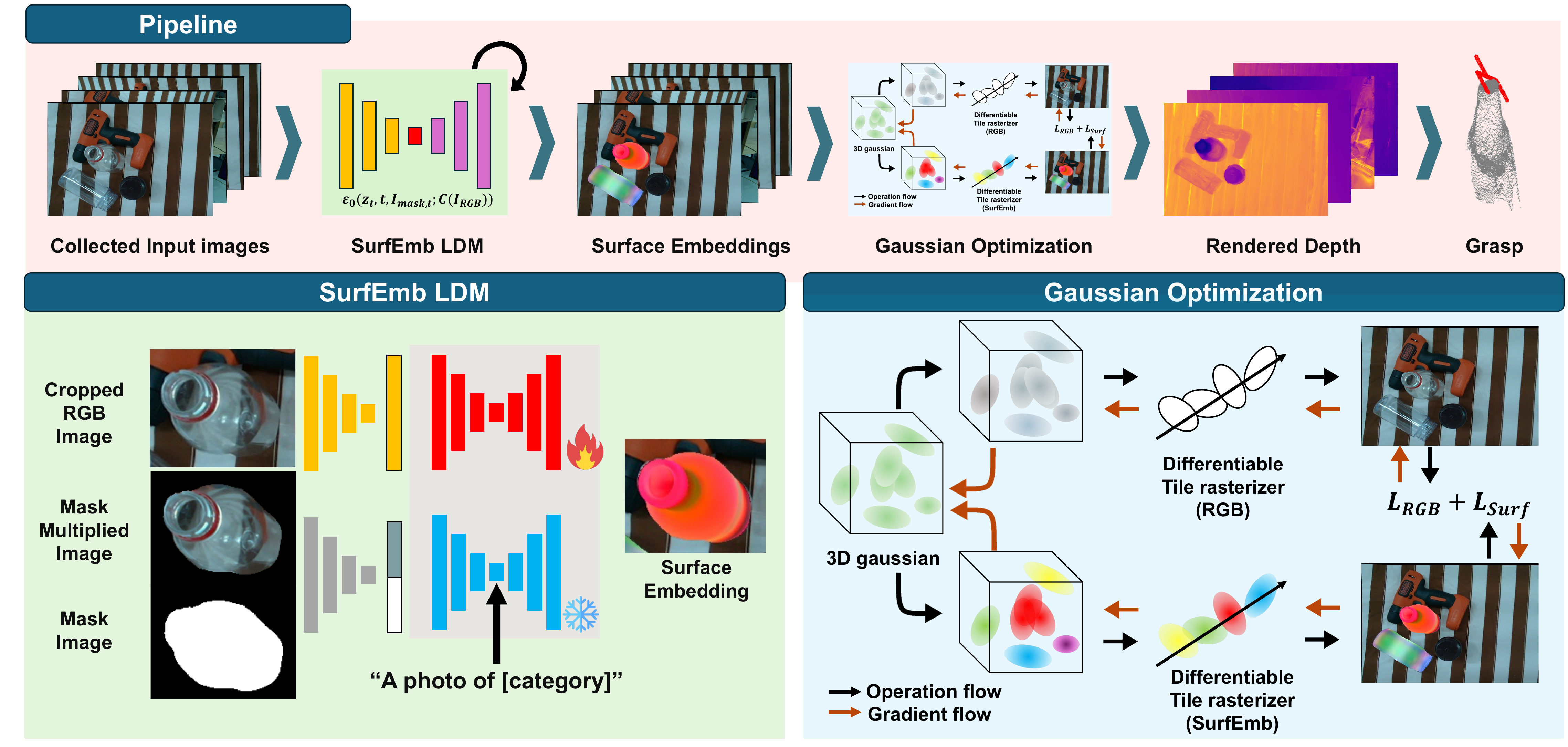}
    \caption{Overview of the TranSplat method for manipulating transparent objects. First, data is collected using a robot manipulator. Next, a latent diffusion model is employed to learn surface embeddings. Then, both surface embeddings and RGB images are used for joint Gaussian optimization. Finally, depth is rendered to enable accurate robotic grasping.}
    \label{fig:method-overview}
    \vspace{-6mm}
\end{figure*}


In our work, we propose TranSplat (\figref{fig:figure1}), a novel method combining the strengths of \ac{3D-GS} \cite{kerbl20233d} and latent diffusion models to improve the reconstruction of transparent objects. TranSplat uses latent diffusion models to extract surface embeddings—continuous surface representations \cite{lee2023nfl, haugaard2022surfemb}—from transparent object features, ensuring consistent representations that are robust to changes in illumination and viewpoint. This reduces noise and artifacts in depth maps. Additionally, TranSplat introduces a jointly-optimized \ac{3D-GS} approach that synthesizes novel views of transparent objects by using both surface embeddings and RGB images. The surface embeddings, acting as surrogate features for non-Lambertian surfaces, prevent the collapse of opacity values and yield accurate depth representation of transparent surfaces. Moreover, employing \ac{3D-GS} instead of NeRF not only speeds up rendering but also enhances depth completion accuracy for transparent objects.

TranSplat demonstrates significant improvements in depth completion accuracy on both synthetic datasets and the real-world TRansPose dataset \cite{kim2024transpose}. We further evaluate its effectiveness in depth estimation by applying it to transparent object manipulation, achieving accurate detection of grasping points. The key contributions of our work include:

\begin{itemize}
    \item \textbf{Diffusion-based Surface Embeddings}: We introduce a novel latent diffusion model specifically designed for transparent objects. This model generates background-agnostic surface embeddings that provide consistent representations of transparent surfaces, regardless of viewpoint and illumination changes. By leveraging surface embeddings, our approach achieves enhanced interframe consistency across consecutive RGB images, improving the overall quality of depth completion.
    \item \textbf{Gaussian Splatting for Transparent object}: We propose an enhanced \ac{3D-GS} method through joint optimization of Gaussian kernels using both RGB images and surface embeddings. This approach effectively captures the surface characteristics of transparent objects, achieving accurate depth reconstruction. We further demonstrate the efficacy of our method through real world grasping of transparent objects.     
    \item \textbf{Open-sourcing Synthetic Dataset}: Our model and the synthetic datasets used for this work will be open-sourced for future development to this field. 
    
\end{itemize}





\section{related work}
\label{sec:relatedwork}

\subsection{Explicit Representation for Robot Manipulation}


In robot manipulation, explicit object representations such as keypoints \cite{mjeon-2022-ral} and object poses have been commonly used, but recent studies suggest that continuous surface representations, like SurfEmb \cite{haugaard2022surfemb}, offer better modeling capabilities, especially for symmetric objects \cite{haugaard2023multi}. SurfEmb facilitates 2D-3D matching by generating dense features from 3D CAD models; however, its reliance on CAD models and the need for separate networks for each object limit its scalability. To address these issues, NeuSurfEmb \cite{milano2024neusurfemb} employs \ac{NeRF} to create large-scale synthetic datasets, enabling dense correspondence matching without CAD models. In our work, we leverage SurfEmb for transparent objects due to its scene-agnostic nature, which ensures consistent representation across consecutive frames, making it effective for dynamic environments.

\subsection{Latent Diffusion for Representation Generation}

With the growing popularity of latent diffusion models for image generation, these models have also demonstrated versatility in various vision tasks, such as depth estimation \cite{Ke2024marigold}, object detection \cite{chen2023diffusiondet}, optical flow \cite{saxena2024surprising}, and visual navigation \cite{sridhar2024nomad}. In robotic manipulation, diffusion models have been utilized to formulate representations for pose estimation. A notable example is 6D-Diff \cite{xu20246ddiff}, which leverages diffusion models to generate keypoint representations, resulting in improved pose estimation accuracy. To the best of our knowledge, our work is the first to employ latent diffusion models to generate explicit representations of transparent objects in the form of SurfEmb.

\subsection{Depth Completion for Grasping Transparent Objects}


Depth completion for transparent objects presents unique challenges that are still being addressed by the research community. Supervised methods rely on paired image-depth data from existing datasets \cite{ichnowski-2022-corl, chen-2022-eccv, kim2024transpose}, but obtaining accurate 3D CAD models for novel objects is difficult. Moreover, achieving visual fidelity in synthetic data and obtaining precise ground truth in real data remain challenging, leading to reduced performance in out-of-domain scenarios and limiting effectiveness in practical applications like robotic grasping.

Recent approaches have used radiance field-based methods \cite{ichnowski-2022-corl, kerr-2022-corl, duisterhof2024residualnerf} for depth completion through 3D scene reconstruction. Although \ac{NeRF}-based methods, including those using \ac{SH} coefficients, have shown promise in handling non-Lambertian surfaces, they struggle with transparent objects due to inconsistencies caused by reflection and refraction. Concurrent methods have tried to mitigate inter-frame inconsistencies by extracting geometry using object masks \cite{chen2023nerrf, ummadisingu2024said}. While these techniques achieve higher surface density through MLP outputs, they often face challenges in maintaining consistency and rely heavily on mask priors, which complicates handling the complexities of transparent objects.
\section{Methods}
As shown in \figref{fig:method-overview}, TranSplat operates in two stages. In the first stage, a latent diffusion model is used to extract surface embeddings from each transparent object in the RGB image, providing a consistent representation of the object across different viewpoints. In the second stage, these surface embeddings, combined with the RGB image, are utilized to render depth and reconstruct 3D scenes through \ac{3D-GS}. 

\subsection{Diffusion-based Surface Embedding Extraction}

To enhance depth completion for transparent objects, TranSplat generates surface embeddings using a latent diffusion model. Inspired by SurfEmb \cite{haugaard2022surfemb}, which effectively captures surface characteristics of various objects, we hypothesize that surface embeddings can provide improved depth completion and a viewpoint-agnostic representation for transparent objects.

To train TranSplat, four data components are required: input RGB image, corresponding mask for transparent object, text condition, and ground truth surface embedding. We trained the model in SurfEmb \cite{haugaard2022surfemb} to generate surface embedding ground truth. However, SurfEmb relies on object-specific networks trained using 3D CAD models, limiting its scalability to real-world scenarios with unknown objects. In our work, we adopt a more generalizable approach where we leverage a category-level training approach, enabling the network to generate similar features for objects within the same category rather than assigning an object specific CAD model. This allows the model to generalize to a wider range of unseen objects, making it more practical for real-world applications. The modified SurfEmb network is used to generate ground truths for training.

\begin{figure}[h]
  \centering
  \begin{subfigure}{0.49\columnwidth}
    \centering
    \includegraphics[width=\textwidth]{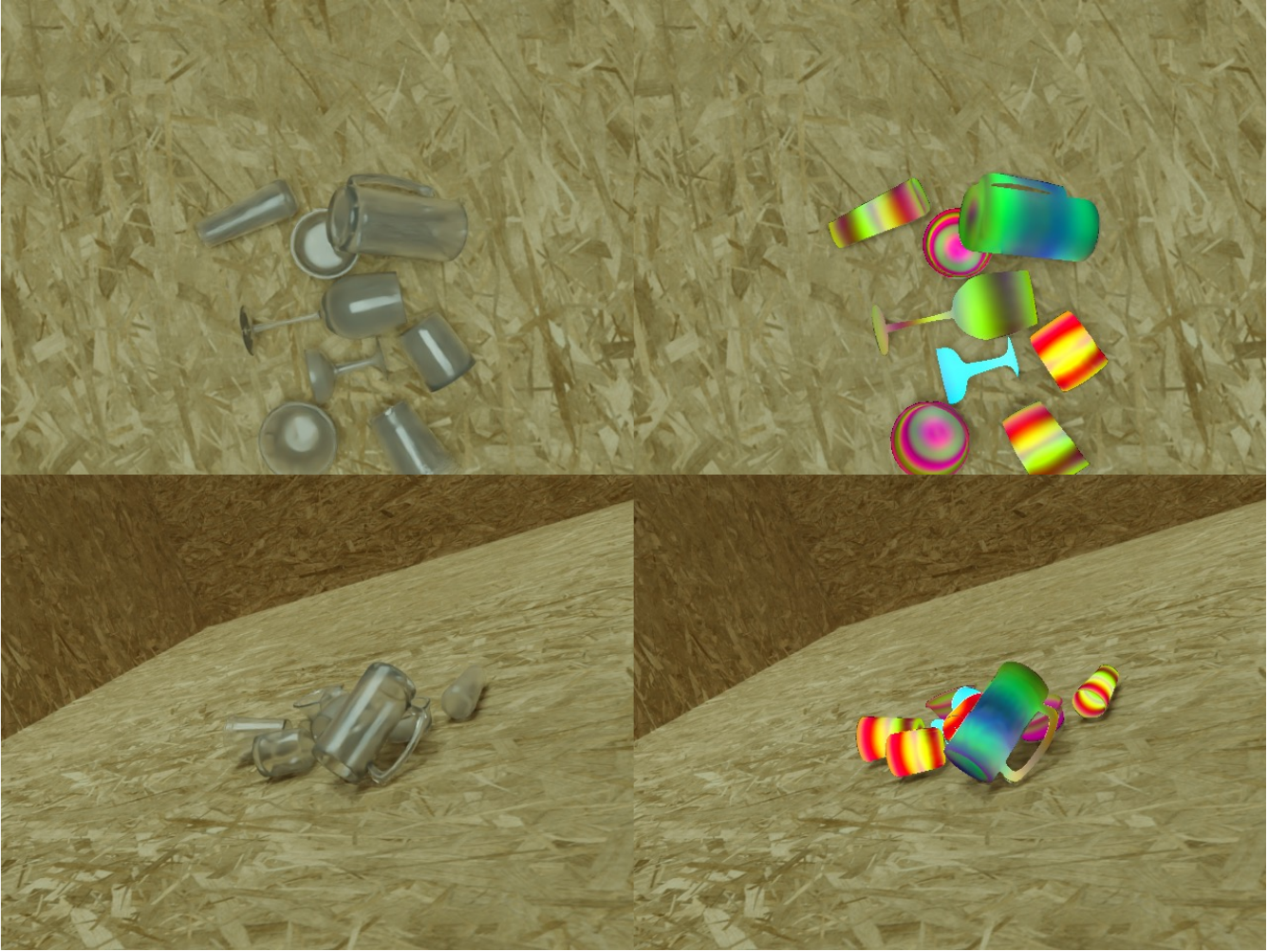}
    \caption{Synthetic unseen object}
    \label{fig:unknown_syn}
  \end{subfigure}
  \begin{subfigure}{0.49\columnwidth}
    \centering
\includegraphics[width=\textwidth]{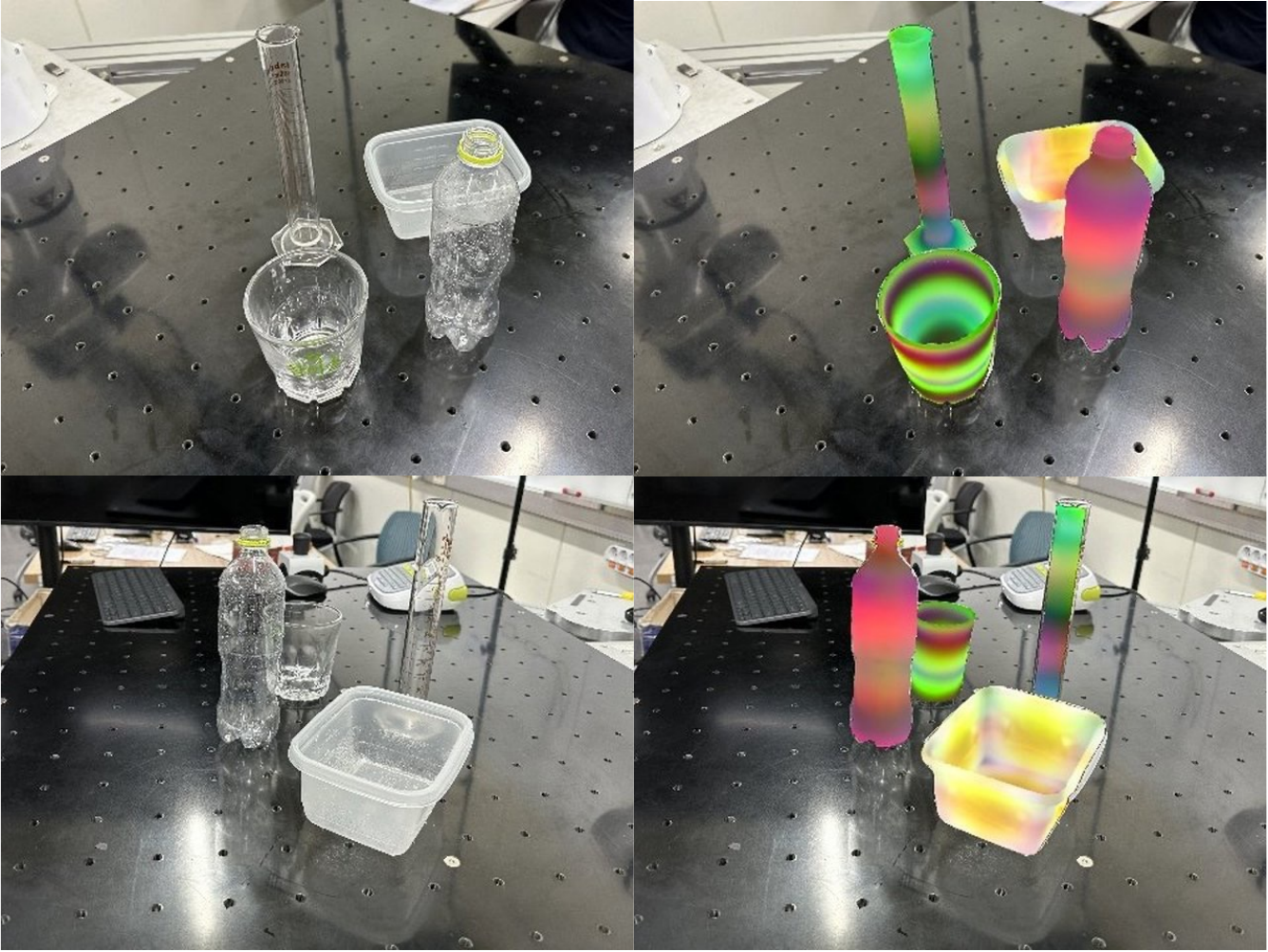}
    \caption{Real world unseen object}
    \label{fig:unknown_real}
  \end{subfigure}
  \caption{Surface embeddings visualization for unseen transparent objects.}
  \label{fig:surfemb-example}
  \vspace{-6mm}
\end{figure}

To extract surface embeddings from RGB images with a latent diffusion model, we concatenate latents generated from the image mask and the mask-multiplied RGB image in the forward process. Text conditioning, consisting of categorical descriptions of objects, is also applied (See \figref{fig:method-overview} green box). Using ControlNet \cite{zhang2023controlnet} architecture, we employ the cropped RGB image as input control. This control helps guide surface embedding generation for specific objects, particularly in scenes with multiple clustered objects. Examples of the generated surface embeddings are shown in \figref{fig:surfemb-example}  

\subsection{Gaussian Splatting for Transparent Objects}

\subsubsection{Color and Depth Rendering for 3D Gaussian Splatting}

To achieve faster rendering speeds than existing \ac{NeRF} models, we use \ac{3D-GS} for depth completion of transparent objects. \ac{3D-GS} represents 3D scenes as a collection of Gaussian distributions, with each Gaussian kernel parameterized by its position, color, size, orientation, and visibility. This approach enables smooth and realistic scene rendering. The color and depth of the rendered scenes are computed using these Gaussian attributes, as shown in \eqref{color_original} and \eqref{depth_original}.
\vspace{-2mm}

\small
\begin{equation}
C = \sum_{j \in N} c_{j} \cdot \alpha_{j} \cdot T_j  \text{, where} \ T_{j} = \prod^{j-1}_{k=1} (1-\alpha_{k})
\label{color_original}
\vspace{-2mm}
\end{equation}

\begin{equation}
    D = \frac{\sum_{j \in N} d_{j} \cdot \alpha_{j} \cdot T_j}{\sum_{j \in N} \alpha_{j} \cdot T_j}
\label{depth_original}
\end{equation}

\normalsize
where $c, d, \alpha, T$ each represents kernel color, kernel depth, opacity, and the accumulated transmittance for the $j_{th}$ observed Gaussian kernel \cite{rendering_eq, yang2024deformable, matsuki2024gaussian}.

\subsubsection{Joint Gaussian Optimization for Transparent Objects}

Applying \ac{3D-GS} to non-Lambertian surfaces, such as transparent objects, often results in low opacity values and reduced $\alpha$ coefficients. Consequently, the Gaussian kernels on transparent surfaces are obstructed during the splatting process, leading to incomplete depth reconstruction. This issue is further exacerbated by varying backgrounds and viewpoints, reducing depth accuracy.

To address this, TranSplat modifies the \ac{3D-GS} rendering process by incorporating surface embedding coefficients. Unlike prior methods that rely solely on rasterizing RGB images, TranSplat rasterizes reconstructed images using the \ac{SH} coefficients for both RGB and surface embeddings. This dual rasterization allows for independent rendering of both RGB images and surface embeddings. The modified rendering equation is demonstrated in  \eqref{color} and \eqref{surf}.




\begin{equation}
C_{RGB} = \sum_{j \in N} c_{RGB, j} \cdot \alpha_{j} \cdot T_j
\label{color}
\vspace{-2mm}
\end{equation}

\begin{equation}
C_{Surf} = \sum_{j \in N} c_{Surf, j} \cdot \alpha_{j} \cdot T_j
\label{surf}
\end{equation}

Moreover, we also reformulate the gaussian optimize loss function to consider images formulated by both RGB and surface embeddings, as shown in \eqref{modified_loss}.
\begin{figure*}[]
    \centering
    \includegraphics[width=0.9\linewidth]{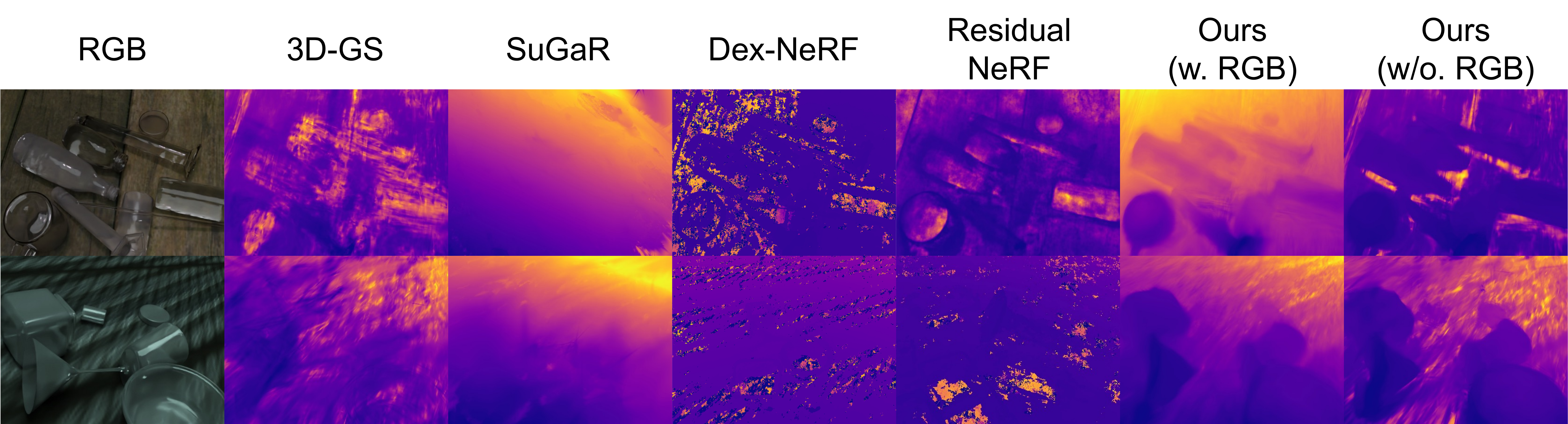}
    \caption{Depth completion results of TRansPose (Top) and ClearPose (Bottom) synthetic sequences.}
    \label{fig:syn}
    \vspace{-2mm}
\end{figure*}


\begin{table*}[ht]
\centering
\caption{Depth completion results for synthetic TRansPose. Best results highlighted in \textbf{bold}; Second best in \underline{underlines}.}
\label{tab:syn-transpose}
\resizebox{\textwidth}{!}{%
\begin{tabular}{ccccccccccccc}
\hline
\multicolumn{13}{c}{Dataset: Synthetic Transpose}\\ \hline
\multicolumn{1}{c|}{Evaluation Metric} & \multicolumn{6}{c|}{MAE $\downarrow$}& \multicolumn{6}{c}{RMSE $\downarrow$}\\ \hline
\multicolumn{1}{c|}{Test Sequences}     & 3D-GS & SuGar  & Dex-NeRF                                              & \begin{tabular}[c]{@{}c@{}}Residual\\ NeRF\end{tabular} & \begin{tabular}[c]{@{}c@{}}Ours\\ (w/o RGB)\end{tabular} & \multicolumn{1}{c|}{\begin{tabular}[c]{@{}c@{}}Ours\\ (w/ RGB)\end{tabular}} & 3D-GS & SuGar  & Dex-NeRF                                              & \begin{tabular}[c]{@{}c@{}}Residual\\ NeRF\end{tabular} & \begin{tabular}[c]{@{}c@{}}Ours\\ (w/o RGB)\end{tabular} & \begin{tabular}[c]{@{}c@{}}Ours\\ (w/ RGB)\end{tabular} \\ \hline
\multicolumn{1}{c|}{1} & \textbf{0.0338} & 0.2720 & 0.0882 & 0.1365 & 0.2809 & \multicolumn{1}{c|}{\underline{0.0406}} & \textbf{0.0478} & 0.4213 & 0.2523 & 0.4130 & 0.4503 & \underline{0.0537}\\
\multicolumn{1}{c|}{2} & 0.1002 & 0.0790 & 0.0644 & \underline{0.0402} & 0.0852 & \multicolumn{1}{c|}{\textbf{0.0272}} & 0.1796 & 0.1601 & 0.2522 & \underline{0.1186} & 0.1690 & \textbf{0.0585}\\
\multicolumn{1}{c|}{3} & 0.0805 & \ding{55} & 0.3423 & 0.1793 & \underline{0.0292} & \multicolumn{1}{c|}{\textbf{0.0199}} & 0.1001 & \ding{55} & 0.6294 & 0.4066 & \underline{0.0581} & \textbf{0.0377}\\
\multicolumn{1}{c|}{5} & 0.1648 & 0.0874 & 0.2839 & \underline{0.0549} & 0.0578 & \multicolumn{1}{c|}{\textbf{0.0280}} & 0.2272 & 0.1523 & 0.8764 & 0.1907 & \underline{0.1358} & \textbf{0.0648}\\
\multicolumn{1}{c|}{6} & 0.0867 & 0.0637 & 0.0641 & \underline{0.0450} & 0.0633 & \multicolumn{1}{c|}{\textbf{0.0266}} & 0.1618 & \underline{0.1296} & 0.2548 & 0.1317 & 0.1508 & \textbf{0.0589}\\
\multicolumn{1}{c|}{7} & 0.1059 & 0.1018 & 0.2660 & 0.1367 & \underline{0.0549} & \multicolumn{1}{c|}{\textbf{0.0275}} & 0.1764 & 0.1723 & 0.5546 & 0.2700 & \underline{0.1389} & \textbf{0.0558}\\
\multicolumn{1}{c|}{8} & 0.2441 & 0.1056 & 0.6620 & 0.3431 & \underline{0.0730} & \multicolumn{1}{c|}{\textbf{0.0357}} & 0.3103 & 0.1841 & 1.4330 & 0.6932 & \underline{0.1474} & \textbf{0.0857}\\
\multicolumn{1}{c|}{9} & 0.2008 & 0.1870 & 0.1595 & 0.3047 & \underline{0.1513} & \multicolumn{1}{c|}{\textbf{0.1111}} & 0.2915 & 0.3024 & 0.3913 & 0.4908 & \underline{0.2483} & \textbf{0.1743}\\
\multicolumn{1}{c|}{10} & 0.1848 & 0.0758 & 0.3870 & 0.0965 & \underline{0.0486} & \multicolumn{1}{c|}{\textbf{0.0255}} & 0.2520 & 0.1460 & 1.0841 & 0.4234 & \underline{0.1140} & \textbf{0.0559} \\
\multicolumn{1}{c|}{11} & 0.0504 & 0.2128 & \textbf{0.0419} & \underline{0.0434} & 0.3103 & \multicolumn{1}{c|}{0.0452} & \underline{0.0762} & 0.3708 & 0.1237 & 0.1301 & 0.4835 & \textbf{0.0673} \\
\multicolumn{1}{c|}{12} & 0.0562 & 0.0921 & \underline{0.0521} & 0.1084 & 0.2008 & \multicolumn{1}{c|}{\textbf{0.0470}} & \underline{0.0903} & 0.1723 & 0.3075 & 0.3747 & 0.3613 & \textbf{0.0779}\\
\multicolumn{1}{c|}{13} & 0.0426 & 0.1340 & \underline{0.0399} & 0.1961 & 0.2777 & \multicolumn{1}{c|}{\textbf{0.0361}} & \underline{0.0719} & 0.3077 & 0.1515 & 0.5154 & 0.4938 & \textbf{0.0661}\\
\multicolumn{1}{c|}{14} & \underline{0.0521} & 0.2654 & 0.2815 & 0.1768 & 0.4307 & \multicolumn{1}{c|}{\textbf{0.0461}} & \underline{0.0883} & 0.4851 & 0.9132 & 0.6196 & 0.6325 & \textbf{0.0809}\\
\multicolumn{1}{c|}{15} & \underline{0.0702} & 0.1097 & 0.6331 & 0.3125 & 0.1553 & \multicolumn{1}{c|}{\textbf{0.0606}} & \underline{0.1599} & 0.1651 & 1.5973 & 0.7698 & 0.3031 & \textbf{0.1250}\\ \hline
\end{tabular}%
}
\vspace{-2mm}
\end{table*}

\begin{table*}[ht]
    \centering
    \caption{Depth completion results for synthetic ClearPose. Best results highlighted in \textbf{bold}; Second best in \underline{underlines}.}
    \label{tab:real-clearpose}
    \resizebox{\textwidth}{!}{%
    \begin{tabular}{ccccccccccccc}
    \hline
    \multicolumn{13}{c}{Dataset: Synthetic ClearPose}\\ \hline
    \multicolumn{1}{c|}{Evaluation Metric} & \multicolumn{6}{c|}{MAE $\downarrow$}& \multicolumn{6}{c}{RMSE $\downarrow$}\\ \hline
    \multicolumn{1}{c|}{Test Sequences}     & 3D-GS & SuGar  & Dex-NeRF & \begin{tabular}[c]{@{}c@{}}Residual\\ NeRF\end{tabular} & \begin{tabular}[c]{@{}c@{}}Ours\\ (w/o RGB)\end{tabular} & \multicolumn{1}{c|}{\begin{tabular}[c]{@{}c@{}}Ours\\ (w/ RGB)\end{tabular}} & 3D-GS & SuGar  & Dex-NeRF & \begin{tabular}[c]{@{}c@{}}Residual\\ NeRF\end{tabular} & \begin{tabular}[c]{@{}c@{}}Ours\\ (w/o RGB)\end{tabular} & \begin{tabular}[c]{@{}c@{}}Ours\\ (w/ RGB)\end{tabular} \\ \hline
    \multicolumn{1}{c|}{1} & 0.2152 & \underline{0.1438} & 0.6038 & 0.5403 & 0.1534 & \multicolumn{1}{c|}{\textbf{0.0519}} & 0.2984 & \underline{0.2604} & 1.4545 & 1.2959 & 0.3159 & \textbf{0.1244}\\
    \multicolumn{1}{c|}{2} & 0.1799 & 0.1483 & 0.2773 & 0.1401 & \underline{0.1376} & \multicolumn{1}{c|}{\textbf{0.0739}} & 0.2246 & \underline{0.2139} & 0.4952 & 0.4732 & 0.2502 & \textbf{0.1979}\\
    \multicolumn{1}{c|}{3} & 0.1835 & \underline{0.0876} & 0.4600 & 0.2405 & 0.1672 & \multicolumn{1}{c|}{\textbf{0.0548}} & 0.3315 & \underline{0.1700} & 0.9566 & 0.7027 & 0.3425 & \textbf{0.1014}\\
    \multicolumn{1}{c|}{4} & 0.1950 & 0.1155 & \textbf{0.0465} & 0.0935 & 0.1871 & \multicolumn{1}{c|}{\underline{0.0858}} & 0.3893 & 0.2868 & \textbf{0.1808} & 0.2300 & 0.3975 & \underline{0.1978}\\
    \multicolumn{1}{c|}{5} & 0.1967 & 0.1725 & 0.0786 & \textbf{0.0335} & 0.1693 & \multicolumn{1}{c|}{\underline{0.0343}} & 0.3419 & 0.3342 & 0.3284 & \underline{0.1141} & 0.3457 & \textbf{0.0622}\\
    \multicolumn{1}{c|}{6} & 0.3300 & 0.3566 & 0.0582 & \underline{0.0406} & 0.3377 & \multicolumn{1}{c|}{\textbf{0.0364}} & 0.5041 & 0.5101 & 0.2471 & \underline{0.1236} & 0.5648 & \textbf{0.0580}\\\hline
    \end{tabular}%
    }
    \vspace{-4mm}
    \end{table*} 

\small
\begin{equation}
    L = \frac{1}{2}L_{RGB} + \frac{1}{2}L_{Surf}
\label{modified_loss}
\vspace{-4mm}
\end{equation}

\begin{equation}
    L_{RGB} = (1-\lambda)|\hat{I}_{RGB}-I_{RGB}| + \lambda\text{D-SSIM}({\hat{I}_{RGB}, I_{RGB}})
\label{RGB_loss}
\vspace{-4mm}
\end{equation}

\begin{equation}
    L_{Surf} = (1-\lambda)|\hat{I}_{Surf}-I_{Surf}| + \lambda \text{D-SSIM}({\hat{I}_{Surf}, I_{Surf}})
\label{Surf_loss}
\end{equation}

\normalsize

where $I_{RGB}$ is the RGB image, $I_{Surf}$ is the surface embeddings image, and $\lambda = 0.2$. This combined loss function optimizes both RGB content and surface features, providing additional supervision to the surfaces of transparent objects. During backward gradient propagation, the Gaussian kernels' mean, covariance, and $\alpha$ values are shared between the RGB and surface embedding images (See \figref{fig:method-overview} blue box). This joint optimization ensures consistent updates of the \ac{SH} coefficients for both representations, allowing the surface embeddings to prevent opacity values, $\alpha$, from collapsing to zero on transparent object surfaces.



\section{experiment}
\label{sec:experiment}
\subsection{Experiment Setup}  
\subsubsection{Datasets}

We evaluated the performance of TranSplat on completing depth from novel rendered views using three datasets: one real-world transparent object dataset with known categories and two synthetic datasets. The first synthetic dataset contains identical objects to those in the real-world dataset, while the second has unseen object models but within the same categories. All synthetic datasets were rendered using BlenderProc \cite{blender}.

For the real-world dataset, we used the TransPose benchmark \cite{kim2024transpose}, which consists of multispectral, multiview sequential images of transparent objects across 20 categories. Each sequence contains 52 and 53 images with corresponding object depth ground truths for training and testing, respectively. For the synthetic datasets, we created two versions: Synthetic TransPose and Synthetic ClearPose. The Synthetic TransPose dataset was rendered using 3D CAD models provided by the real TransPose dataset, matching both the categories and specific objects. In contrast, the Synthetic ClearPose dataset features different object models within the same categories, designed to test TranSplat's performance on unseen objects. Both synthetic datasets contain 100 sequential images per sequence with ground truth depths for training and testing.
\begin{figure*}[]
    \centering
    \includegraphics[width=0.85\linewidth]{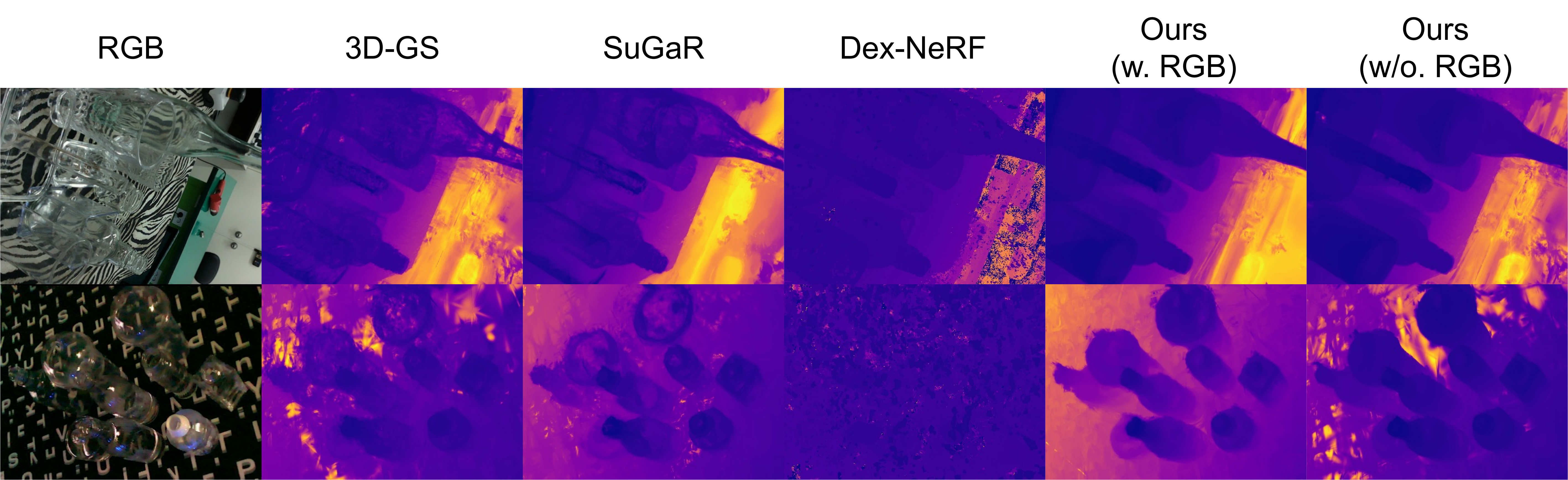}
    \caption{Depth completion results of TRansPose test sequence 7 and 26. }
    \label{fig:real}
    \vspace{-2mm}
\end{figure*}

\begin{table*}[ht]
    \centering
    \caption{Depth completion results for Real TRansPose. Best results highlighted in \textbf{bold}; Second best in \underline{underlines}.}
    \label{tab:real-transpose}
    \resizebox{0.9\textwidth}{!}{%
    \begin{tabular}{ccccccccccc}
    \hline
    \multicolumn{11}{c}{Dataset: Real TRansPose}\\ \hline
    \multicolumn{1}{c|}{Evaluation Metric} & \multicolumn{5}{c|}{MAE $\downarrow$}& \multicolumn{5}{c}{RMSE $\downarrow$}\\ \hline
    \multicolumn{1}{c|}{Test Sequences}     & 3D-GS & SuGar  & Dex-NeRF & \begin{tabular}[c]{@{}c@{}}Ours\\ (w/o RGB)\end{tabular} & \multicolumn{1}{c|}{\begin{tabular}[c]{@{}c@{}}Ours\\ (w/ RGB)\end{tabular}} & 3D-GS & SuGar  & Dex-NeRF & \begin{tabular}[c]{@{}c@{}}Ours\\ (w/o RGB)\end{tabular} & \begin{tabular}[c]{@{}c@{}}Ours\\ (w/ RGB)\end{tabular} \\ \hline
    \multicolumn{1}{c|}{1} & 0.0334 & 0.0377 & 0.0345 & \underline{0.0175} &  \multicolumn{1}{c|}{\textbf{0.0139}} &0.0588 & 0.0644 & 0.0891 & \underline{0.0454} & \textbf{0.0232}\\
    \multicolumn{1}{c|}{2} & 0.0809 & 0.0930 & 0.0958 & \textbf{0.0181} & \multicolumn{1}{c|}{\underline{0.0270}} & 0.1377 & 0.1727 & 0.2494 & \textbf{0.0513} & \underline{0.0561}\\
    \multicolumn{1}{c|}{3} & 0.0606 & 0.0666 & 0.0615 & \textbf{0.0225} & \multicolumn{1}{c|}{\underline{0.0246}} & 0.0994 & 0.1076 & 0.2244 & \underline{0.0561} & \textbf{0.0534}\\
    \multicolumn{1}{c|}{4} & 0.0373 & 0.0445 & 0.0385 & \underline{0.0261} & \multicolumn{1}{c|}{\textbf{0.0189}} & 0.0757 & 0.0905 & 0.1485 & \underline{0.0712} & \textbf{0.0694}\\
    \multicolumn{1}{c|}{5} & 0.0467 & 0.0544 & 0.0471 & \textbf{0.0165} & \multicolumn{1}{c|}{\underline{0.0223}} & 0.0773 & 0.0957 & 0.1959 & \underline{0.0539} & \textbf{0.0527}\\
    \multicolumn{1}{c|}{6} & 0.1067 & 0.1196 & 0.0929 & \textbf{0.0162} & \multicolumn{1}{c|}{\underline{0.0367}} & 0.1799 & 0.2067 & 0.2764 & \textbf{0.0464} & \underline{0.1011}\\
    \multicolumn{1}{c|}{7} & 0.0500 & 0.0520 & 0.0398 & \underline{0.0291} & \multicolumn{1}{c|}{\textbf{0.0185}} & 0.0881 & 0.0927 & 0.1296 & \underline{0.0875} & \textbf{0.0478}\\
    \multicolumn{1}{c|}{8} & 0.0744 & 0.0944 & 0.0920 & \textbf{0.0182} & \multicolumn{1}{c|}{\underline{0.0264}} & 0.1300 & 0.1827 & 0.2904 & \textbf{0.0505} & \underline{0.0580}\\
    \multicolumn{1}{c|}{9} & 0.0566 & 0.0673 & 0.1807 & \textbf{0.0229} & \multicolumn{1}{c|}{\textbf{0.0229}} & 0.0976 & 0.1293 & 0.5911 & \underline{0.0704} & \textbf{0.0702}\\
    \multicolumn{1}{c|}{10} & 0.0592 & 0.0607 & 0.0550 & \textbf{0.0245} & \multicolumn{1}{c|}{\underline{0.0257}} & 0.0997 & 0.1191 & 0.2284 & \textbf{0.0584} & \underline{0.0691}\\
    \multicolumn{1}{c|}{11} & 0.0944 & 0.0931 & 0.0626 & \textbf{0.0278} & \multicolumn{1}{c|}{\underline{0.0368}} & 0.1745 & 0.1957 & 0.1906 & \textbf{0.0762} & \underline{0.1036}\\
    \multicolumn{1}{c|}{12} & 0.0404 & 0.0448 & 0.0473 & \underline{0.0225} & \multicolumn{1}{c|}{\textbf{0.0189}} & 0.0951 & 0.1231 & 0.1360 & \underline{0.0719} & \textbf{0.0587}\\
    \multicolumn{1}{c|}{13} & 0.0431 & 0.0442 & 0.0507 & \underline{0.0292} & \multicolumn{1}{c|}{\textbf{0.0172}} & 0.0910 & 0.0946 & 0.1492 & \underline{0.0855} & \textbf{0.0441}\\
    \multicolumn{1}{c|}{25} & 0.0726 & 0.0618 & 0.1829 & \underline{0.0549} & \multicolumn{1}{c|}{\textbf{0.0191}} & \underline{0.1250} & 0.1394 & 0.4837 & 0.1279 & \textbf{0.0406}\\
    \multicolumn{1}{c|}{26} & 0.0404 & 0.0480 & 0.1831 & \textbf{0.0141} & \multicolumn{1}{c|}{\underline{0.0191}} & 0.0682 & 0.1043 & 0.3041 & \textbf{0.0388} & \underline{0.0499}\\ \hline
    \end{tabular}%
    }
    \vspace{-2mm}
    \end{table*}
\subsubsection{Implementation Details}

For TranSplat, we trained both the latent diffusion-based surface embeddings extractor and the \ac{3D-GS} for neural rendering. Built on ControlNet \cite{zhang2023controlnet}, we froze the latent diffusion UNet and kept the ControlNet counterpart trainable. Training was performed on $256 \times 256$ images from both real and synthetic TRansPose datasets, with a batch size of 32 and a learning rate of 1.0e-6, using the AdamW optimizer with a cosine scheduler. For the \ac{3D-GS}, we followed the settings described in 3D-GS \cite{kerbl20233d}. All models were trained on four Nvidia A6000 GPUs. Further details on the training configurations are available on our project page.

\subsubsection{Baselines}

We used four models as baselines for our evaluations: DexNeRF \cite{ichnowski-2022-corl} and Residual-NeRF \cite{duisterhof2024residualnerf}, which are recent models for novel view completion of transparent objects, as well as \ac{3D-GS} \cite{kerbl20233d} and SuGaR \cite{guedon2024sugar}, an object surface-aligned \ac{3D-GS} model. For TranSplat, we present two variations: one with RGB image input control (w/ RGB) and one without it (w/o RGB). To assess depth completion performance across the baselines and TranSplat, we used mean average error (MAE) and root mean squared error (RMSE) to compare the absolute depth measurements between the ground truths and the rendered views.

\subsection{Evaluation on Synthetic Datasets}

\subsubsection{Evaluation on Synthetic TRansPose}


As shown in \tabref{tab:syn-transpose}, TranSplat achieves the best depth completion performance, outperforming all baseline models in terms of both MAE and RMSE across all sequences. Unlike other models that rely directly on raw images of transparent objects as inputs, TranSplat leverages surface embeddings as an alternative representation, leading to a significant improvement in depth completion. Additionally, TranSplat does not require extensive volume density tuning or separate residual background images—both impractical for robotics applications. \ac{3D-GS} methods often yield near-zero opacity values due to the non-Lambertian nature of transparent objects. Specifically, in synthetic TRansPose dataset sequence 3, the overall darkness of the dataset causes the opacities of Gaussians for transparent objects to converge to zero, resulting in the failure of SuGaR during the pruning step as no Gaussians remain after pruning for further optimization. In contrast, TranSplat's surface embeddings serve as surrogate features that accurately estimate opacity values on transparent surfaces.

The qualitative results, presented in \figref{fig:syn}, further support these findings. Most baseline methods struggle to capture the depth along the edges of transparent objects. Even methods like Dex-NeRF, which do capture some edge details, display incomplete depth with holes around transparent surfaces. This is due to conventional \ac{NeRF}-based methods neglecting the opacity values for transparent surfaces. In contrast, by incorporating the unique properties of transparent objects and supplementing 3D gaussian optimization with surface embeddings, TranSplat achieves complete and dense reconstructions around transparent object surfaces.

\subsubsection{Evaluation on Unseen Synthetic}

We evaluated TranSplat's robustness to unseen objects using the Synthetic ClearPose dataset, as shown in Table \ref{tab:real-clearpose}. Consistent with previous findings, TranSplat achieves the best depth completion performance across all test sequences on average. While it slightly underperforms compared to Residual-NeRF on sequence 5 and Dex-NeRF on sequence 4, the performance gaps are minimal. We attribute the lower performance on sequence 4 to the fact that the objects in this sequence differ significantly from the CAD models used to train the latent diffusion model in the TransPose dataset. Despite this, TranSplat demonstrates the highest robustness to unseen objects, proving its effectiveness across different categories rather than being object-specific. Qualitatively, as shown in \figref{fig:syn}, other methods fail to accurately render the depth images of transparent objects, whereas TranSplat consistently succeeds.

\subsection{Evaluation on Real-world Dataset}


As shown in \tabref{tab:real-transpose}, TranSplat outperforms baseline models in depth completion when evaluated on the real-world TRansPose dataset. However, unlike the consistent results seen in the synthetic dataset evaluation, the best performance in the real-world dataset alternates between TranSplat models with and without RGB images as conditioning input. In the synthetic dataset, incorporating RGB images as input control to the latent diffusion model consistently leads to better results. This is because RGB images provide valuable context and guidance, allowing the model to better attend to transparent objects during depth completion, as illustrated in \figref{fig:real}. In contrast, in the real-world scenario, using RGB input does not always improve performance. The RGB images in real-world settings are often affected by factors like poor lighting, image sensor noise, and severe occlusion between objects, which can negatively impact the model's effectiveness when using RGB conditioning. Additionally, it is important to note that we were unable to evaluate TranSplat against Residual NeRF for the real-world dataset because Residual NeRF requires background images, which were not provided in the TransPose dataset.

\begin{table}[]
\centering
\caption{Depth completion results on reducing the number of images. Best results highlighted in \textbf{bold}; Second best in \underline{underlines}.}
\begin{adjustbox}{width=1\linewidth}
\begin{tabular}{c|ccc|ccc}
\hline
Evaluation Metric        & \multicolumn{3}{c|}{MAE 
$\downarrow$}                                                                       & \multicolumn{3}{c}{RMSE $\downarrow$}                                                                      \\ \hline
\multirow{2}{*}{Dataset} & Synthetic                           & Real                          & Synthetic                            & Synthetic                           & Real                          & Synthetic                           \\
                         & \multicolumn{1}{l}{TRansPose} & \multicolumn{1}{l}{TRansPose} & \multicolumn{1}{l|}{ClearPose} & \multicolumn{1}{l}{TRansPose} & \multicolumn{1}{l}{TRansPose} & \multicolumn{1}{l}{ClearPose} \\ \hline
3D-GS               & 0.1066                        & 0.0598                        & 0.2167                         & 0.1618                        & 0.1065                        & 0.3483                        \\
SuGar                    & 0.1327                        & 0.0655                        & 0.1707                         & 0.2362                        & 0.1279                        & 0.2959                        \\
Dex-nerf                 & 0.2347                        & 0.0839                        & 0.2541                         & 0.6291                        & 0.2461                        & 0.6104                        \\
Residual-nerf            & 0.1503                        & \ding{55}                             & 0.1814                         & 0.3910                        & \ding{55}                             & 0.4899                        \\
Ours(w/o RGB)                  & 0.1740                        & \underline{0.0240}                        & 0.1921                         & 0.2903                        & \underline{0.0661}                        & 0.3694                        \\
Ours(w/ RGB)           & \textbf{0.0406}                        & \textbf{0.0232}                        & \textbf{0.0562}                         & \textbf{0.0757}                        & \textbf{0.0599}                        & 0.1236                        \\
Ours(w/ RGB, 1/2)                & 0.0449                        & 0.0360                        & \underline{0.0563}                         & 0.0820                        & 0.0875                        & \underline{0.1185}                        \\
Ours(w/ RGB, 1/4)                & \underline{0.0418}                        & 0.1120                        & 0.0646                         & \underline{0.0766}                        & 0.2154                        & \textbf{0.1161}                        \\ \hline
\end{tabular}
\end{adjustbox}
\label{ablation}
\vspace{-5mm}
\end{table}

\subsection{Computational Efficiency Analysis}

A practical approach to reducing computation time in robotics applications is to decrease the number of images used for rendering. This is particularly relevant when the sensor's sampling rate does not support high frame rates, resulting in sparse image outputs. To evaluate this, we analyzed the accuracy-efficiency trade-off of TranSplat by varying the number of images used. As shown in Table \ref{ablation}, reducing the number of images leads to a slight decrease in performance. However, despite this minor degradation, TranSplat still outperforms other baseline models that use more images. Although TranSplat's use of diffusion models can result in slower inference times with a full sequence of images, reducing the number of rendered images can significantly enhance computational efficiency with only a minor drop in accuracy. This also simplifies the system overhead by allowing for a lower sampling rate of visual sensors while maintaining competitive performance.

\subsection{Extension to Transparent Object Grasping}

\begin{figure}[!t]
    \centering
    \includegraphics[width=1\linewidth]{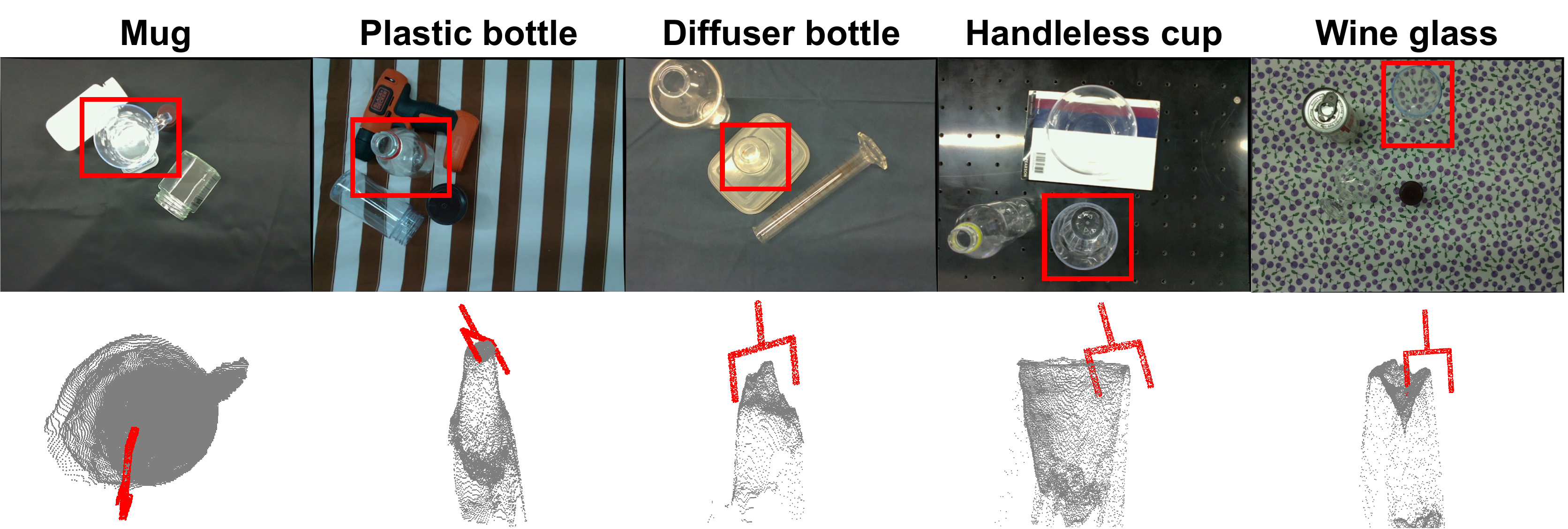}
    \caption{The first column is RGB images. The second column shows the grasp planning locations from Graspnet. }
    \label{fig:grasp_result}
    \vspace{-5mm}
\end{figure}

To explore the feasibility of extending TranSplat for robot manipulation through grasping, we tested its performance using a commercial robot arm. We used the Franka Emika Panda to capture a series of RGB images of unseen transparent objects, as shown in the \figref{fig:grasp_result}. These images were then used to generate input point clouds by combining RGB data with the corresponding depth rendered by TranSplat. To determine the grasping points, we employed the pretrained GraspNet model \cite{fang2020graspnet}, which generates grasp points from input point clouds. The point clouds were created using RGB images and the depth outputs generated by TranSplat. As shown in \figref{fig:grasp_result}, GraspNet successfully identifies valid grasping points using the depth rendered from TranSplat. This demonstrates that the depth produced by TranSplat provides accurate depth information that can be effectively used for robotic manipulation of transparent objects. We encourage readers to refer to the supplementary materials for more details.

\section{Conclusion}
\label{sec:conclusion}

Accurately capturing the depth of transparent objects remains a significant challenge for conventional depth sensors, which often struggle with transparency. Existing methods using radiance field-based techniques attempt to address this by rendering depth from novel views, but they often fail to handle transparent surfaces adequately, leading to incomplete depth renderings. In this work, we introduced TranSplat, which overcomes this limitation by incorporating surface embeddings generated through latent diffusion models. TranSplat consistently outperforms existing methods in accurately capturing the depth of transparent objects in both synthetic and real-world datasets, including practical applications in robot grasping tasks. For future work, we plan to enhance TranSplat by estimating uncertainties in the input RGB images used for conditioning, further improving its robustness and applicability.


\newpage


\balance
\small
\bibliographystyle{IEEEtranN} 
\bibliography{string-short,references}

\end{document}